\documentclass[11pt,a4paper]{article}
\usepackage[hyperref]{naaclhlt2019}
\usepackage{times}
\usepackage{latexsym}
\usepackage{comment}
\usepackage{graphicx}
\usepackage{booktabs}
\usepackage{subcaption}
\usepackage{enumitem}

\usepackage{amssymb}
\usepackage{pifont}

\usepackage{multirow}
\usepackage{sidecap}
\usepackage{amsthm}
\usepackage{amsmath}
\usepackage{amssymb}
\usepackage{amsfonts}
\usepackage{verbatim} 
\usepackage{url}
\usepackage{pifont}
\usepackage{eqparbox}
\usepackage{arydshln} 
\usepackage{bigstrut}
\usepackage[linewidth=1pt]{mdframed}
\usepackage{framed}
\usepackage{float}
\usepackage{url}
\usepackage{cancel}

\usepackage{array}
\newcolumntype{L}[1]{>{\raggedright\let\newline\\\arraybackslash\hspace{0pt}}m{#1}}
\newcolumntype{C}[1]{>{\centering\let\newline\\\arraybackslash\hspace{0pt}}m{#1}}
\newcolumntype{R}[1]{>{\raggedleft\let\newline\\\arraybackslash\hspace{0pt}}m{#1}}

\aclfinalcopy 

\definecolor{DarkRed}{RGB}{130,25,0}
\definecolor{DarkGreen}{RGB}{30,130,30}

\newcommand{\todo}[1]{{\color{purple} [TODO: {#1}]}}
\newcommand{\cmark}{\textcolor{DarkGreen}{\ding{51}}}
\newcommand{\xmark}{\textcolor{red}{\ding{55}}}%

\newcommand{\ignore}[1]{}
\newcommand{\claim}{\emph{claim}}
\newcommand{\perspective}{\emph{perspective}}
\newcommand{\evidence}{\emph{evidence}}

\newcommand{\datasetname}{\textsc{\textcolor[wave]{390}{P}\textcolor[wave]{415}{e}\textcolor[wave]{440}{r}\textcolor[wave]{465}{s}\textcolor[wave]{485}{p}\textcolor[wave]{525}{e}\textcolor[wave]{535}{c}\textcolor[wave]{595}{t}\textcolor[wave]{610}{r}\textcolor[wave]{635}{u}\textcolor[wave]{660}{m}}}

\newcommand{\bert}{\textsc{BERT}}

\newcommand{\evidencePool}{\mathcal{U}^e}
\newcommand{\perspectivePool}{\mathcal{U}^p}

\newcommand{\set}[1]{\{#1\}}
\newcommand{\abs}[1]{\left|#1\right|}
\newcommand{\llbrac}{[\![}
\newcommand{\rrbrac}{]\!]}
\newcommand{\eqcls}[1]{\llbrac #1\rrbrac}
\newcommand{\oneE}[1]{\mathbf{1}\set{#1}}

\title{
\vspace*{-0.5in}
{{\small \hfill NAACL'19}\\
\vspace*{.25in}}
Seeing Things from a Different Angle: \\ Discovering Diverse Perspectives about Claims }

\author{Sihao Chen, Daniel Khashabi,  Wenpeng Yin,  Chris Callison-Burch, Dan Roth \\
Department of Computer and Information Science, University of Pennsylvania \\
  {\tt \small \{sihaoc,danielkh,wenpeng,ccb,danroth\}@cis.upenn.edu} 
}

\date{}

\usepackage{xcolor}
\usepackage{pgffor}

\begin{document}
\maketitle
\begin{abstract}

One key consequence of the information revolution is a significant increase and a contamination of our information supply. The practice of fact-checking won't suffice to eliminate the biases in text data we observe, as the degree of factuality alone does not determine whether biases exist in the spectrum of opinions visible to us. To better understand controversial issues, one needs to view them from a diverse yet comprehensive set of {\em perspectives}.

For example, there are many ways to respond to a \emph{claim} such as \emph{``animals should have lawful rights''}, and these responses form a spectrum of perspectives, each with a {\em stance} relative to this claim and, ideally, with evidence supporting it. 
Inherently, this is a natural language understanding task, and we propose to address it as such. Specifically, we propose the task of \emph{substantiated perspective discovery} where, given a \emph{claim}, a system is expected to discover a \emph{diverse} set of \emph{well-corroborated} \emph{perspectives} that take a \emph{stance} with respect to the claim. Each perspective should be substantiated by \emph{evidence} paragraphs which summarize pertinent results and facts. 

We construct \datasetname, a dataset of claims, perspectives and evidence, making use of online debate websites to create the initial data collection, and augmenting it using search engines in order to expand and diversify our dataset. We use crowdsourcing to filter out noise and ensure high-quality data.  
Our dataset contains 1$k$ claims, accompanied by pools of 
10$k$ and 
8$k$ perspective sentences and evidence paragraphs, respectively. We provide a thorough analysis of the dataset to highlight key underlying language understanding challenges, and show that human baselines across multiple subtasks far outperform machine baselines built upon state-of-the-art NLP techniques. 
  This poses a challenge and an opportunity for the NLP community to address.
\end{abstract}


\section{Introduction}
Understanding most nontrivial {\em claim}s requires insights from various \perspective s.
Today, we make use of search engines or recommendation systems to retrieve information relevant to a claim, but this process carries multiple forms of \emph{bias}. In particular, they are optimized relative to the claim (query) presented, and the popularity of the relevant documents returned, rather than with respect to the diversity of the \perspective s presented in them or whether they are supported by evidence.

\begin{figure}
    \centering
    \includegraphics[scale=0.37,trim=1.2cm 0.5cm 0cm 0.0cm, clip=false]{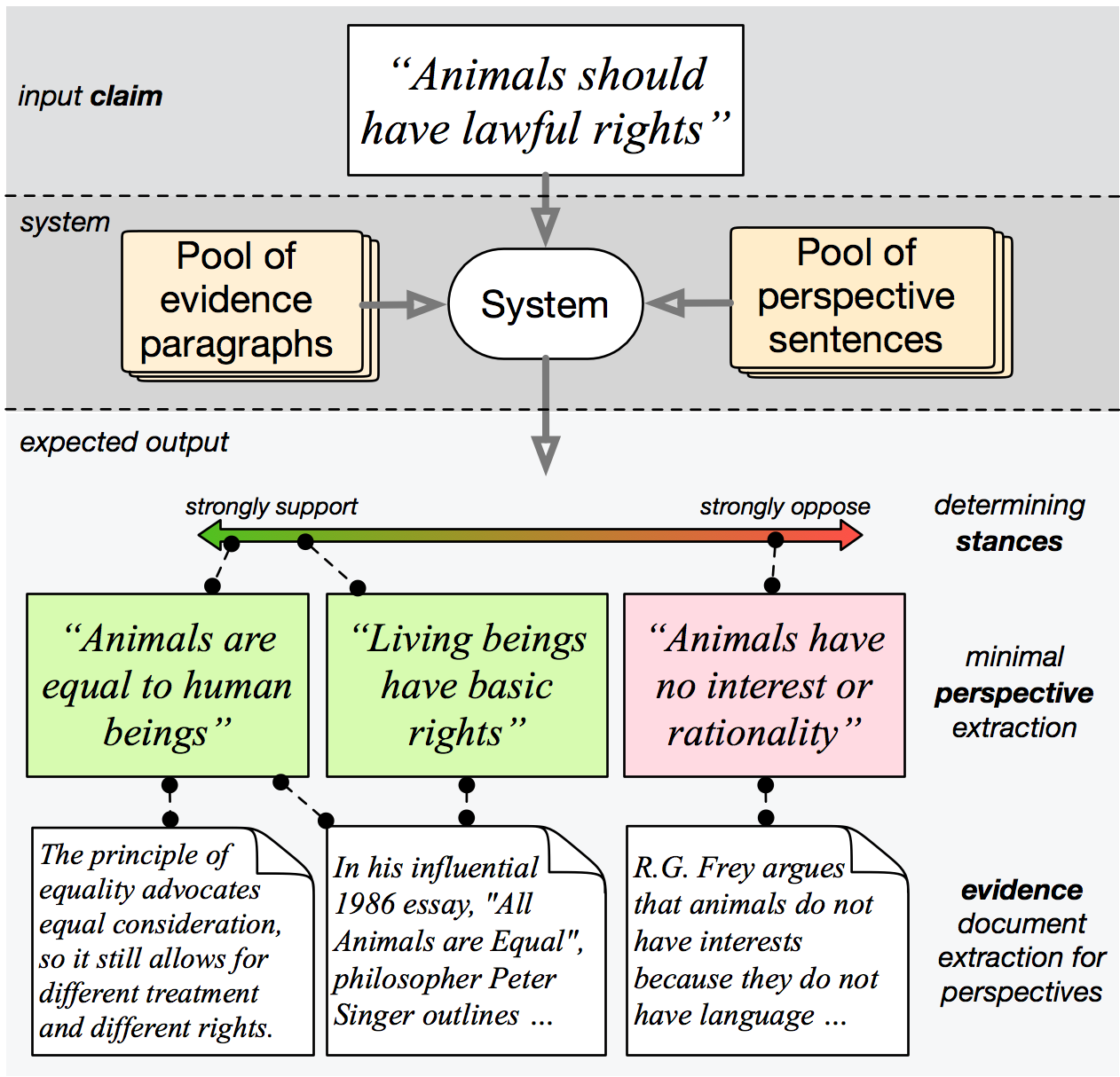}
    \caption{Given a \emph{claim}, a hypothetical system is expected to discover various \emph{perspectives} that are substantiated with \emph{evidence} and their \emph{stance} with respect to the claim. }
    \label{fig:example:intro}
\end{figure}
In this paper, we explore an approach to mitigating this
\emph{selection bias}~\cite{H79} when studying (disputed) claims. 
Consider the \emph{claim} shown in Figure \ref{fig:example:intro}: \emph{``animals should have lawful rights.''}
One might compare the biological similarities/differences between humans and other animals to support/oppose the claim. 
Alternatively, one can base an argument on morality and rationality of animals, or lack thereof.
Each of these arguments, which we refer to as \perspective s throughout the paper, is an opinion, possibly conditional, in support of a given \emph{claim} or against it.  
A \perspective\, thus constitutes a particular attitude towards a given \emph{claim}. 

Natural language understanding is at the heart of developing an ability to identify diverse perspectives for claims. In this work, we propose and study a setting that would facilitate discovering \emph{diverse perspectives} and their supporting evidence with respect to a given \emph{claim}. Our goal is to identify and formulate the key NLP challenges underlying this task, and develop a dataset that would allow a systematic study of these challenges.   
For example, for the claim in Figure~\ref{fig:example:intro}, multiple (non-redundant) perspectives should be retrieved from a pool of perspectives; one of them is \emph{``animals have no 
interest 
or rationality''}, a \perspective\ that should be identified as taking an \emph{opposing} stance with respect to the \emph{claim}. 
Each \emph{perspective} should also be well-supported by \emph{evidence} found in  a  pool of potential pieces of evidence.
While it might be impractical to provide an exhaustive spectrum of ideas with respect to a \emph{claim}, 
presenting
a small but diverse set of \emph{perspectives} could be an important 
step towards addressing the \emph{selection bias} problem. Moreover,
it would be impractical to develop an exhaustive pool of evidence for all perspectives, from a diverse set of credible sources. We are not attempting to do that. We aim at formulating the core NLP problems, and developing a dataset that will facilitate studying these problems from the NLP angle, realizing that using the outcomes of this research in practice requires addressing issues such as trustworthiness~\cite{PasternackRo10a,PasternackRo13} and possibly others. Inherently, our objective requires understanding the relations between \perspective s and \claim s, the nuances in the meaning of various \perspective s in the context of \claim s, and relations between perspectives and evidence. This, we argue, can be done with a diverse enough, but not exhaustive, dataset. And it can be done without attending to the legitimacy and credibility of sources contributing evidence, an important problem but orthogonal to the one studied here.
To facilitate the research towards developing solutions to such challenging issues, we propose \datasetname, a dataset of \emph{claims}, \emph{perspectives} and \emph{evidence} paragraphs.
For a given \emph{claim} and pools of \emph{perspectives} and \emph{evidence paragraphs}, a hypothetical system is expected to select the relevant perspectives and their supporting paragraphs. 


Our dataset contains 907 claims, 11,164 perspectives and 8,092 evidence paragraphs. In constructing it, we use online debate websites as our initial seed data, and augment it with search data and paraphrases to make it richer and more challenging. We make extensive use of crowdsourcing to increase the quality of the data and clean it from annotation noise.   





The contributions of this paper are as follows: 
\begin{itemize}[noitemsep,leftmargin=0.4cm]
    \item To facilitate making progress towards the problem of \emph{substantiated perspective discovery}, we create a high-quality dataset for this task.\footnote{https://github.com/CogComp/perspectrum }
    \item We identify and formulate multiple NLP tasks that are at the core of addressing the \emph{substantiated perspective discovery} problem. We show that humans can achieve high scores on these tasks. 
    \item 
    We develop competitive baseline systems for each sub-task, using state-of-the-art techniques.
\end{itemize}

\definecolor{lightgreen}{RGB}{200, 255, 200}
\definecolor{lightred}{RGB}{255, 200, 200}

\begin{figure}
    \centering
    \includegraphics[scale=0.38, trim=0cm 0.5cm 0cm 0.0cm, clip=false]{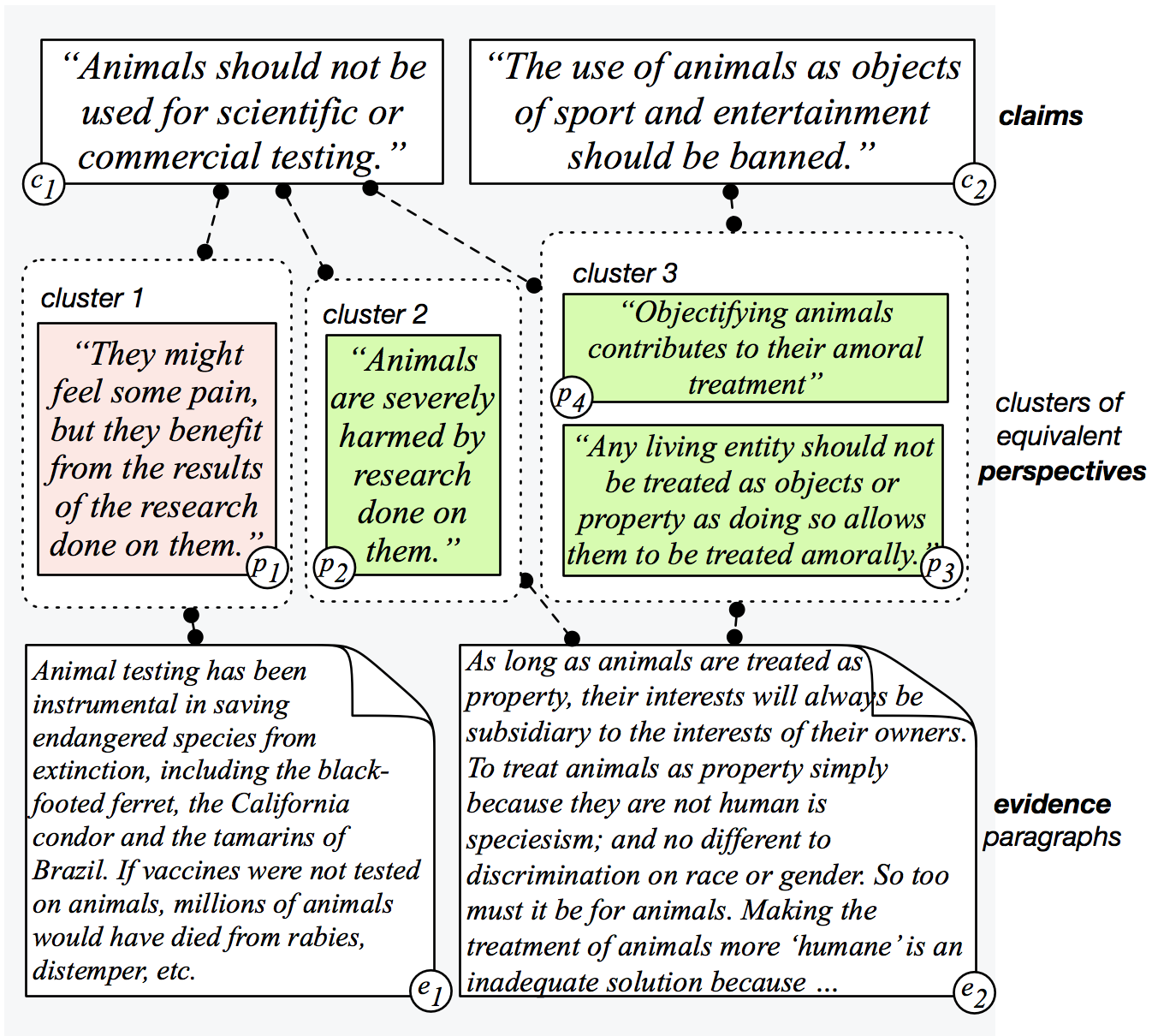}
    \caption{
    Depiction of a few claims, their \emph{perspectives} and  evidences from \datasetname. 
    The \emph{supporting} and \emph{opposing} perspectives are indicated with \colorbox{lightgreen}{green} and \colorbox{lightred}{red} colors, respectively. 
    }
    \label{fig:clusters}
\end{figure}


\section{Design Principles and Challenges}
In this section we provide a closer look into the challenge and propose a collection of tasks that move us closer to \emph{substantiated perspective discovery}. 
To clarify our description we use to following notation. 
Let $c$ indicate a target claim of interest (for example, the claims $c_1$ and $c_2$ in Figure~\ref{fig:clusters}). 
Each claim $c$ is addressed by a collection of perspectives $\set{p}$ that are grouped into clusters of \emph{equivalent} perspectives. Additionally, each perspective $p$ is supported, relative to $c$, by at least one evidence paragraph $e$, denoted $e\vDash p|c$. 

Creating systems that would address our challenge in its full glory requires solving the following interdependent tasks: \\ 
\noindent    
\emph{Determination of argue-worthy claims: } not every claim requires an in-depth discussion of perspectives. For a system to be practical, it needs to be equipped with understanding argumentative structures~\cite{PalauMo09} in order to 
discern disputed claims from those with straightforward responses. We set aside this problem in this work and assume that all the inputs to the systems are discussion-worthy claims. 
    
\noindent
\emph{Discovery of pertinent perspectives:} a system is expected to recognize argumentative sentences~\cite{CabrioVi12} that directly address the points raised in the disputed claim. 
For example, while the perspectives in Figure~\ref{fig:clusters} are topically related to the claims, $p_1, p_2$ do not directly address the focus of claim $c_2$ (i.e., \emph{``use of animals'' in ``entertainment''}). 

\noindent    
\emph{Perspective equivalence: }
a system is expected to extract a \emph{minimal} and \emph{diverse} set of perspectives. This requires the ability to discover equivalent perspectives $p, p'$, with respect to a claim $c$: $p | c \approx p^{'} | c$. For instance, $p_3$ and $p_4$ are equivalent in the context of $c_2$; however, they might not be equivalent with respect to any other claim. 
The conditional nature of perspective equivalence differentiates it from the \emph{paraphrasing} task~\cite{BannardCa05}. 

\noindent  
\emph{Stance classification of perspectives:} a system is supposed to assess the stances of the perspectives with respect to the given claim (supporting, opposing, etc.) ~\cite{HasanNg14b}. 
    
\noindent   
\emph{Substantiating the perspectives:} 
a system is expected to find valid evidence paragraph(s) in support of each perspective. 
Conceptually, this is similar to the well-studied problem of textual entailment~\cite{DRSZ13} 
except that here the entailment decisions depend on the choice of claims. 

\vspace{0.5cm} 

\section{Related Work}
\paragraph{Claim verification.}
The task of \emph{fact verification} or \emph{fact-checking} focuses on the assessment of the truthfulness of a claim, given evidence \cite{VlachosRi14,MitraGi15,STVB16,Wang17,NBESMZAKM18,HPSCCMG18,KRST18,APMa18}.
These tasks are highly related to the task of textual-entailment that has been extensively studied in the field~\cite{BCDG08,DRSZ13,KhotSaCl18}. Some recent work study jointly the problem of identifying evidence and verifying that it supports the claim~\cite{YinRo18b}.

Our problem structure encompasses the \emph{fact verification} problem (as verification of \emph{perspectives} from \emph{evidence}; Figure~\ref{fig:example:intro}). 

\begin{table*}[]
    \centering
    \small 
    \begin{tabular}{c C{1.7cm} C{1.7cm} C{1.7cm} C{1.7cm}}
        \toprule
        Dataset & Stance Classification & Evidence Verification &  Human Verified & Open Domain  \\
         \cmidrule(lr){1-1}  \cmidrule(lr){2-2}  \cmidrule(lr){3-3}  \cmidrule(lr){4-4}  \cmidrule(lr){5-5} 
         \datasetname\ (this work) & \cmark  & \cmark  & \cmark & \cmark  \\ 
         FEVER~\cite{TVCM18} & \xmark & \cmark & \cmark & \cmark  \\ 
         \cite{WPKAPQDMBS17} & \cmark & \cmark & \xmark & \cmark  \\
        LIAR~\cite{Wang17} & \xmark & \cmark & \cmark & \cmark  \\
        \cite{VlachosRi14} & \xmark & \cmark & \cmark & \cmark  \\
         \cite{HasanNg14b} &  \cmark & \xmark & \cmark & \xmark   \\
         \bottomrule 
    \end{tabular}
    \caption{Comparison of \datasetname\ to a few notable datasets in the field.
    }
    \label{tab:comparisons}
\end{table*}

\paragraph{Stance classification.}
Stance classification aims at detecting phrases that \emph{support} or \emph{oppose} a given claim. The problem has gained significant attention in the recent years; to note a few important ones, \newcite{HasanNg14b} create a dataset of dataset text snippets, annotated with ``reasons'' (similar 
to \emph{perspectives} in this work) and stances (whether they support or oppose the claim). Unlike this work, our pool of the relevant ``reasons'' is not restricted. \newcite{FerreiraVl16} create a dataset of rumors (claims) coupled with news headlines and their stances. There are a few other works that fall in this category~\cite{Boltuvzic14,ParkCa14,RDPKAS15,SwansonEcWa15,MKSZC16,SobhaniInZh17,BBDSS17}. 
Our approach here is closely related to existing work in this direction, as stance classification is part of the problem studied here. 




\paragraph{Argumentation.}
There is a rich literature on \emph{formalizing} argumentative structures from free text. 
There are a few theoretical works that lay the ground work to characterizing units of arguments and argument-inducing inference \cite{Teufelot99,Toulmin03,Freeman11}. 

Others have studied the problem of extracting argumentative structures from free-form text; for example, 
\newcite{PalauMo09,KWKHS16,ACKWS17} studied elements of arguments and the internal relations between them. 
\newcite{FengHi11} classified an input into one of the argument schemes. 
\newcite{HabernalGu17} provided a large corpus annotated  with argument units. 
\newcite{CabrioVi18} provide a thorough survey the recent work in this direction. A few other works studied other aspects of argumentative structures~\cite{CabrioVi12,KWKHS16,LippiTo16,ZHHL17,StabGu17}.

A few recent works use a similar conceptual design that involves a \emph{claim}, \emph{perspectives} and \emph{evidence}.These works are either too small due to the high cost of construction~\cite{APLHLRGS14} or too noisy because of the way they are crawled from online resources~\cite{WPKAPQDMBS17,HuaWa17}. Our work makes use of both online content and of crowdsourcing, in order to construct a sizable and high-quality dataset. 

\section{The \datasetname\ Dataset}
\subsection{Dataset construction}
\label{sec:construction}
In this section we describe 
a multi-step process, constructed with 
detailed analysis, substantial refinements and multiple pilots studies. 

We use crowdsourcing to annotate different aspects of the dataset. We used Amazon Mechanical Turk (AMT) for our annotations, restricting the task to workers in five English-speaking
countries (USA, UK, Canada, New Zealand, and Australia), more than 1000 finished HITs and at least a 95\% acceptance rate. 
To ensure the diversity of responses, we do not require additional qualifications or demographic information from our annotators.

For any of the annotations steps described below, the users are guided to an external platform where they first read the instructions and try a verification step to make sure they have understood the instructions. Only after successful completion are they allowed to start the annotation tasks. 

Throughout our annotations, it is our aim to make sure that the workers are responding objectively to the tasks (as opposed to using their  personal opinions or preferences).
The screen-shots of the annotation interfaces for each step are included in the Appendix (Section~\ref{sec:supp:screenshots}). 

In the steps outlined below, we filter out a subset of the data with low rater--rater agreement $\rho$ (see Appendix~\ref{sec:agreement}). In certain steps,
we use an information retrieval (IR) system\footnote{\url{www.elastic.co}} to generate the best candidates for the task at hand.

\paragraph{Step 1: The initial data collection.}
We start by crawling the content of a few notable debating websites: {\tt \small idebate.com, debatewise.org, procon.org}. 
This yields $\sim1k$ claims, $\sim8k$ perspectives and $\sim8k$ evidence paragraphs (for complete statistics, see Table~\ref{tab:seed_data} in the Appendix). This data is significantly noisy and lacks the structure we would like. In the following steps we explain how we denoise it and augment it with additional data. 

\paragraph{Step 2a: Perspective verification. }
For each perspective we verify that it is a complete English sentence, with a clear stance with respect to the given claim. 
For a fixed pair of \emph{claim} and \emph{perspective}, we ask the crowd-workers to label the perspective with one of the five categories of \emph{support}, \emph{oppose}, \emph{mildly-support}, \emph{mildly-oppose}, or \emph{not a valid perspective}. The reason that we ask for two levels of intensity 
is to 
distinguish \emph{mild} or \emph{conditional} arguments from those that express \emph{stronger} positions. 

Every 
10 claims (and their relevant perspectives) are bundled to form a HIT. Three independent annotators solve a HIT, and each gets paid \$1.5-2 per HIT. 
To get rid of the ambiguous/noisy perspectives we measure rater-rater agreement on the resulting data and retain only the subset which has a significant agreement of $\rho \geq 0.5$. 
To account for minor disagreements in the intensity of perspective stances, before measuring any notion of agreement, we collapse the five labels into three labels, by collapsing \emph{mildly-support} and \emph{mildly-oppose} into \emph{support} and \emph{oppose}, respectively. 

To assess the quality of these annotations, 
two of the authors independently annotate a 
 random subset of instances in the previous step (328 perspectives for 10 claims). Afterwards, the differences were adjudicated. 
We measure the accuracy adjudicated results with AMT annotations to estimate the quality of our annotation. This results in an accuracy of 94\%, which shows high-agreement with the crowdsourced annotations.  




\paragraph{Step 2b:  Perspective paraphrases. }
To enrich the 
ways the perspectives are phrased, we crowdsource paraphrases of our 
perspectives. We ask annotators to generate two paraphrases for each of the 15 perspectives in each HIT, for a reward of \$1.50. 

Subsequently, we perform another round of crowdsourcing to verify the generated paraphrases. We create HITs of 24 candidate paraphrases to be verified, with a reward of \$1. Overall, this process gives us $\sim4.5$ paraphrased perspectives. The collected paraphrases form  clusters of equivalent perspectives, which we refine further in the later steps. 


\paragraph{Step 2c:  Web perspectives.} 
In order to ensure that our dataset contains more realistic sentences, we use web search to augment our pool of perspectives with additional sentences that are topically related to what we already have. 
Specifically, we use Bing search to extract sentences that are similar to our current pool of perspectives, 
by querying ``claim+perspective''.
We create a pool of relevant web sentences and use an IR system (introduced earlier) to retrieve the 10 most similar sentences. 
These candidate perspectives are annotated using (similar to step 2a) and only those that were agreed upon are retained. 


\paragraph{Step 2d:  Final perspective trimming.} 
In a final round of annotation  for perspectives, an expert annotator went over all the claims in order to verify that all the equivalent perspectives are clustered together. 
Subsequently, the expert annotator went over the most similar claim-pairs (and their perspectives), in order to annotate the missing perspectives shared between the two claims. 
To cut the space of claim pairs, the annotation was done on the top 350 most similar claim pairs retrieved by the IR system. 

\paragraph{Step 3: Evidence verification. }
The goal of this step is to decide whether a given evidence paragraph provides enough substantiations for a perspective or not. Performing these annotations exhaustively for any perspective-evidence pair is not possible. Instead, we make use of a retrieval system to annotate only the relevant pairs. In particular, we create an index of all the perspectives retained from \emph{step 2a}. For a given evidence paragraph, we retrieve the top relevant perspectives. We ask the annotators to note whether a given evidence paragraph \emph{supports} a given perspective or not. 
Each HIT contains a 20 evidence paragraphs and their top 8 relevant candidate perspectives. Each HIT is paid \$$1$ and annotated by at least 4 independent annotators. 



In order to assess the quality of our annotations, a random subset of instances 
(4 evidence-perspective pairs) are annotated by two independent authors and the differences are adjudicated. 
We measure the accuracy of our adjudicated labels versus AMT labels, resulting in 87.7\%. This indicates the high quality of the crowdsourced data. 

\begin{table}[]
    \small 
    \centering
    \resizebox{\linewidth}{!}{
    \begin{tabular}{clc}
        \toprule
            Category & \multicolumn{1}{c}{Statistic}  & Value \\ 
        \cmidrule(lr){1-1} \cmidrule(lr){2-2} \cmidrule(lr){3-3}
            \multirow{4}{*}{Claims} & \# of claims (step 1) &  907\\
            & avg. claim length (tokens) & 8.9\\ 
            & median claims length (tokens) & 8\\ 
            & max claim length (tokens) & 30\\ 
            & min claim length (tokens) & 3\\ 
            \hline 
            \multirow{5}{*}{Perspectives} & \# of perspectives & 11,164\\
            &     \multicolumn{1}{l}{\hspace{0.5cm} Debate websites (step 1)} & 4,230 \\
            &     \multicolumn{1}{l}{\hspace{0.5cm} Perspective paraphrase (step 2b)} & 4,507 \\
            &     \multicolumn{1}{l}{\hspace{0.5cm} Web (step 2c)} & 2,427 \\
             & \# of perspectives with stances & 5,095\\
             & \# of ``support'' perspectives & 2,627\\
             & \# of ``opposing'' perspectives & 2,468\\
            & avg size of perspective clusters & 2.3 \\
            & avg length of perspectives (tokens) & 11.9 \\
            \hline 
            \multirow{2}{*}{Evidences}  & \# of total evidences (step 1) & 8,092 \\
            & avg length of evidences (tokens) & 168 \\
        \bottomrule
    \end{tabular}
    }
    \caption{A summary of \datasetname~ statistics}
    \label{tab:statistics}
\end{table}

\subsection{Statistics on the dataset}
\label{sec:statistics}
We now provide a brief summary of \datasetname.
The dataset contains about $1k$ claims with a significant length diversity (Table~\ref{tab:statistics}). Additionally, the dataset comes with $\sim 12k$ perspectives, most of which were generated through paraphrasing (step 2b). The perspectives which convey the same point with respect to a claim are grouped into clusters. On average, each cluster has a size of $2.3$ which shows that, on average, many perspectives have equivalents. More granular details are available in Table~\ref{tab:statistics}. 

\begin{figure}
    \centering
    \includegraphics[scale=0.24]{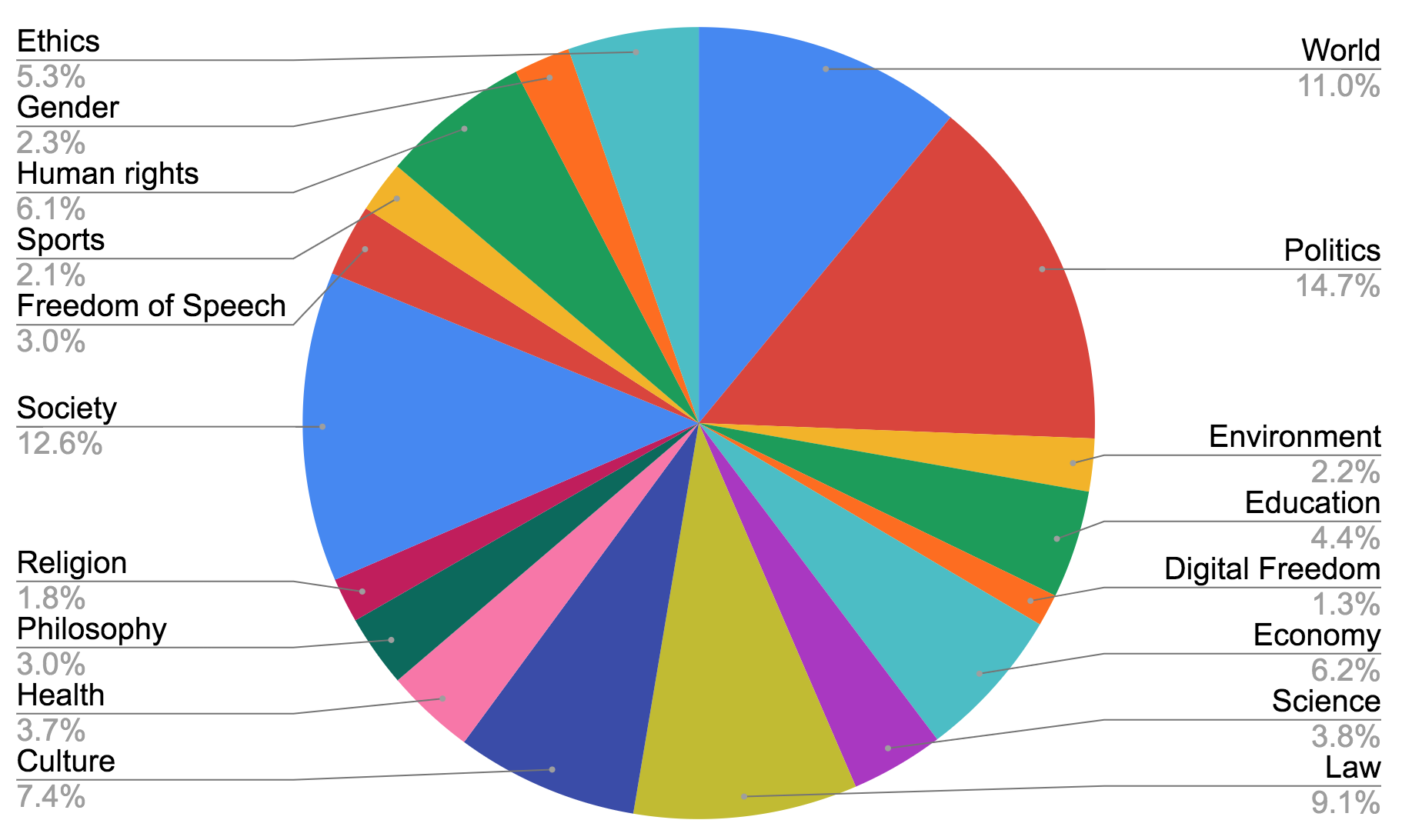}
    \caption{Distribution of claim topics.}
    \label{fig:topic_distribution}
\end{figure}

\begin{figure*}
    \centering
    \includegraphics[scale=0.32]{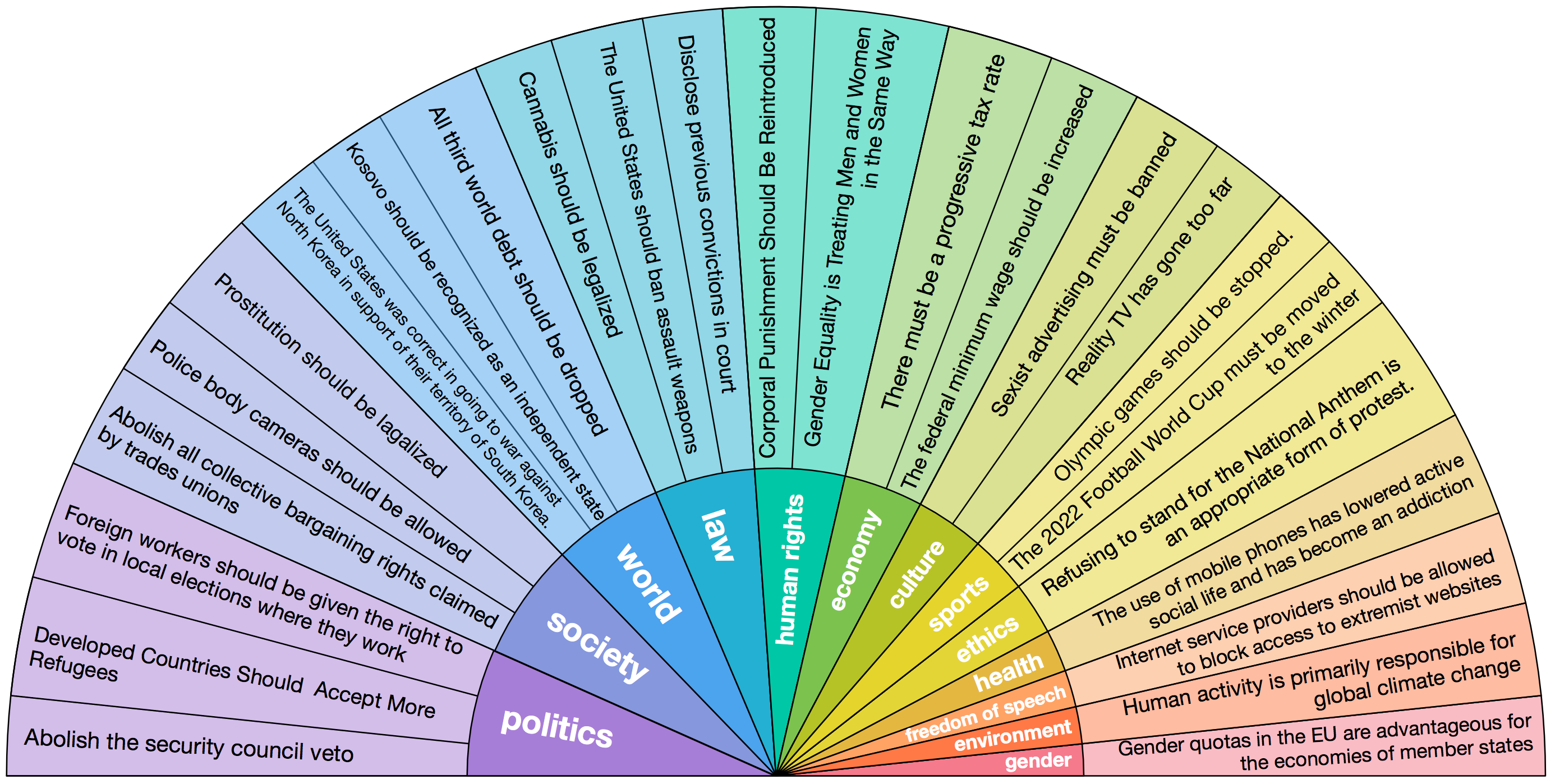}
    \caption{Visualization of the major topics and sample claims in each category.}
    \label{fig:topics}
\end{figure*}

To better understand the topical breakdown of claims in the dataset, we crowdsource the set of ``topics'' associated with each \emph{claim} (e.g., \emph{Law, Ethics, etc}.) We observe that, as expected, the three topics of \emph{Politics, World, and Society} have the biggest portions (Figure~\ref{fig:topic_distribution}). Additionally, the included claims touch upon 10+ different topics. Figure~\ref{fig:topics} depicts a few popular categories and sampled questions from each.

\subsection{Required skills}
\begin{figure}
    \centering
    \includegraphics[scale=0.19,trim=0cm 0cm 0cm 0cm]{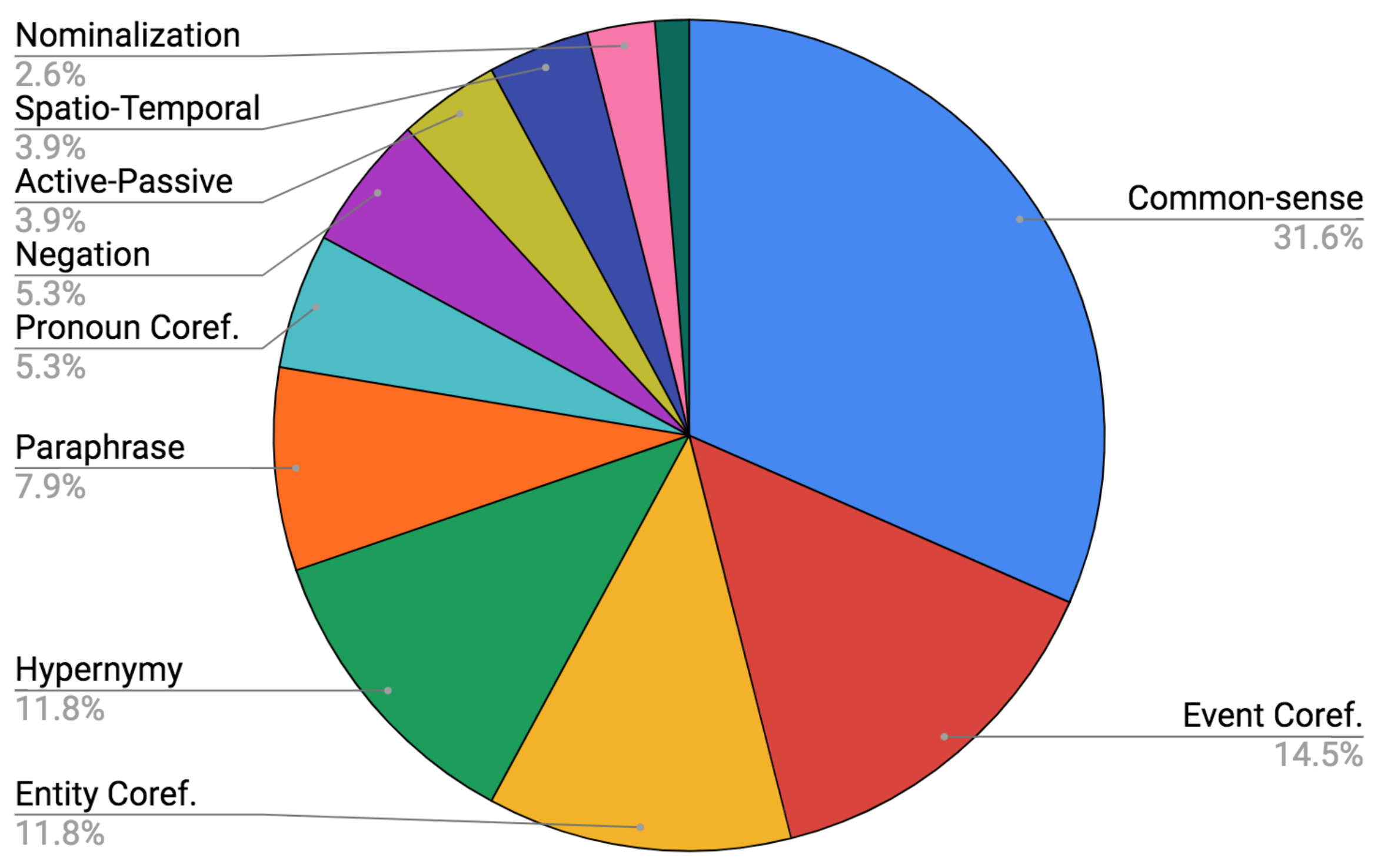}
    \caption{The set of reasoning abilities required to solve the stance classification task. }
    \label{fig:reasoning-categories}
\end{figure}

We perform a closer investigation of the abilities required to solve the stance classification task.
One of the authors went through a random subset of claim-perspectives pairs and annotated each with the abilities required in determining their stances labels. 
We follow the common definitions used in prior work~\cite{SugawaraYoAi17,KCRUR18}. 
The result of this annotation is depicted in Figure~\ref{fig:reasoning-categories}. As can be seen, the problem requires understanding of 
\emph{common-sense}, 
i.e., an understanding that is commonly shared among humans and rarely gets explicitly mentioned in the text. Additionally, the task requires various types of \emph{coreference} understanding, such as \emph{event coreference} and \emph{entity coreference}. 

\section{Empirical Analysis}
\label{sec:analysis}
In this section we provide empirical analysis to address the tasks.
We create a split of 60\%/15\%/25\% of the data train/dev/test. 
In order to make sure our baselines are not overfitting to the keywords of each topic (the ``topic'' annotation from Section~\ref{sec:statistics}), we make sure to have claims with the same topic fall into the same split. 

For simplicity, we define a notation which we will extensively use for the rest 
of this paper. 
The clusters of equivalent perspectives are denoted as $\eqcls{p}$, given a representative member $p$. 
Let $P(c)$ denote the collection of relevant perspectives to a claim $c$, which is the union of all the equivalent perspectives participating in the claim: $\set{ \eqcls{p_i} }_{i}$. 
Let $E(\eqcls{p}) = E(p) = \bigcup_{i} e_i$ denote the set of evidence documents lending support to a perspective $p$. 
Additionally, denote the two pools of perspectives and evidence with  $\perspectivePool$ and  $\evidencePool$, respectively. 

\subsection{Systems}
We make use of the following systems in our evaluation: 

    \paragraph{IR} (Information Retrieval). 
    This baseline has been successfully used for related tasks like Question Answering~\cite{CEKSTTK16}. We create two versions of this baseline: one with the pool of perspectives $\perspectivePool$ and one with the pool of evidences $\evidencePool$. We use this system to retrieve a ranked list of best matching perspective/evidence from the corresponding index.

\paragraph{BERT} (Contextual representations). A recent state-of-the-art contextualized representation~\cite{DCLT18}. 
This system has been shown to be effective on a broad range of natural language understanding tasks. 

\paragraph{Human Performance.}
Human performance provides us with an estimate of the best achievable results on datasets. 
We use human annotators to measure human performance for each task. We randomly sample 10 claims from the test set, and instruct two expert annotators to solve each of T1 to T4.

\subsection{Evaluation metrics.}
We perform evaluations on four different subtasks in our dataset. In all  of the following evaluations, the systems are given the two pools of perspectives $\perspectivePool$ and evidences $\evidencePool$.

    \paragraph{T1: Perspective extraction.} A system is expected to return the collection of mutually disjoint perspectives with respect to a given claim. Let $\hat{P}(c)$ be the set of output perspectives. Define the precision and recall as  
    $
    \text{Pre}(c) = \frac{\sum_{\hat{p} \in \hat{P}(c)} \oneE{ \exists p, s.t. \hat{p} \in \eqcls{p} } }{ \abs{\hat{P}(c)} }
    $
    and 
    $
    \text{Rec}(c) = \frac{\sum_{\hat{p} \in \hat{P}(c)} \oneE{ \exists p, s.t. \hat{p} \in \eqcls{p} } }{ \abs{P(c)} }
    $ respectively. To calculate dataset metrics, the aforementioned per-claim metrics are averaged across all the claims in the test set. 
    
    \paragraph{T2: Perspective stance classification.} Given a claim, a system is expected to label every perspective in $P(c)$ with one of two labels \emph{support} or \emph{oppose}. We use the well-established  definitions of precision-recall for this binary classification task. 
    
    \paragraph{T3: Perspective equivalence.} A system is expected to decide whether two given perspectives are equivalent or not, with respect to a given claim. We evaluate this task in a way similar to a clustering problem. 
    For a pair of perspectives $p_1, p_2 \in P(c)$, a system predicts whether the two are in the same cluster or not. The ground-truth is whether there is a cluster which contains both of the perspectives or not: $\exists \tilde{p}\; s.t
    .\;  \tilde{p} \in P(c) \wedge p_1, p_2 \in \eqcls{\tilde{p}} $. We use this pairwise definition for all the pairs in $P(c)\times P(c)$, for any claim $c$ in the test set. 
    
    \paragraph{T4: Extraction of supporting evidences.} Given a perspective $p$, we expect a system to return all the evidence $\set{e_i}$ from the pool of evidence $\evidencePool$. Let $\hat{E}(p)$ and $E(p)$ be the predicted and gold evidence for a perspective $p$. Define macro-precision and macro-recall as $\text{Pre}(p) = \frac{ \abs{\hat{E}(p) \cap E(p) } }{ \abs{\hat{E}(p)}  } $ and $\text{Rec}(p) = \frac{ \abs{\hat{E}(p) \cap E(p) } }{ \abs{E(p)}  } $, respectively. The metrics are averaged across all the perspectives $p$ participating in the test set. 
\paragraph{T5: 
Overall
performance.}
The goal is to get estimates of the overall performance of the systems. Instead of creating a complex measure that would take all the aspects into account, we approximate the overall performance by multiplying the disjoint measures in $T1$, $T2$ and $T4$. 
While this gives an estimate on the overall quality, it ignores the pipeline structure of the task (e.g., the propagation of the errors throughout the pipeline).
We note that the task of $T3$ (perspective equivalence) is indirectly being measured within $T1$. 
Furthermore, since we do not report an IR performance on $T2$, we use the ``always supp'' baseline instead to estimate an overall performance for IR.

\subsection{Results}

\subsubsection{Minimal perspective extraction (T1)}
\label{sec:exp1:minimal}
Table~\ref{tab:results} shows a summary of the experimental results. 
To measure the performance of the IR system, we use the index containing $\perspectivePool$. Given each claim, we query the top $k$ perspectives, ranked according to their retrieval scores. We tune $k$ on our development set and report the results on the test section according to the tuned parameter. We use IR results as 
candidates for other solvers (including humans). For this task, IR  with top-15 candidates yields $>$90\% recall (for the PR-curve, see Figure~\ref{fig:pr-curves} in the Appendix). In order to train \bert\ on this task, we use the IR candidates as the training instances. We then tune a threshold  on the dev data to select the top relevant perspectives. In order to measure human performance, we create an interface where two human annotators see IR top-$k$ and select a \emph{minimal} set of perspectives (i.e., no two equivalent perspectives). 

\subsubsection{Perspective stance classification (T2)}
\label{sec:exp2:stance}
We measure the quality of perspective stance classification, where the input is a claim-perspective pair, mapped to { \set{support, oppose}}. The candidate inputs are generated on the collection of perspectives $P(c)$ relevant to a claim $c$. To have an understanding of a lower bound for the metric, we measure the quality of an { always-support} baseline. We measure the performance of \bert\ on this task as well, which is about 20\% below human performance. This might be because this task requires a deep understanding of \emph{commonsense} knowledge/reasoning (as indicated earlier in Section~\ref{sec:analysis}). 
Since a retrieval system is unlikely to distinguish perspectives with different stances, we do not report the IR performance for this task. 

\subsubsection{Perspective equivalence (T3)}
\label{sec:exp3:equivalent}
We create instances in the form of $(p_1, p_2, c)$ where $p_1, p_2 \in P(c)$. The expected label is whether the two perspectives belong to the same equivalence class or not. In the experiments, we observe that BERT has a significant performance gain of $\sim 36\%$  over the IR baseline. Meanwhile, this system is behind human performance by a margin of $\sim 20\%$.


\subsubsection{Extraction of supporting evidence (T4)}
\label{sec:exp4:evidence}
We evaluate the systems on the extraction of items from the pool of evidences $\evidencePool$, given a \claim-\perspective\ pair. 
To measure the performance of the IR system working with 
the index containing $\evidencePool$ we issue a query containing the concatenation of a perspective-claim pair. 
Given the sorted results (according to their retrieval confidence score), we select the top candidates using a threshold parameter tuned on the dev set. 
We also use the IR system's candidates (top-60) for other baselines. 
This set of candidates yields a  $>$85\% recall (for the PR-curve, see Figure~\ref{fig:pr-curves} in the Appendix). 
We train \bert\ system to map each (gold) \claim-\perspective\ pair to its corresponding \evidence\ paragraph(s). 
Since each evidence paragraph could be long (hence hard to feed into \bert), we split each evidence paragraph into sliding windows of 3 sentences. 
For each \claim-\perspective\ pair, we use all 3-sentences windows of gold evidence paragraphs as positive examples, and rest of the IR candidates as negative examples. 
In the run-time, if a certain percentage (tuned on the dev set) of the sentences from a given evidence paragraph are predicted as positive by \bert, we consider the whole evidence as positive (i.e. it supports a given \perspective).  
  
Overall, the performances on this task are lower, which could probably be expected, considering the length of the evidence paragraphs. Similar to the previous scenarios, the \bert\ solver has a significant gain over a trivial baseline, while standing behind human with a significant margin. 

\begin{table}[]
    \small 
    \centering
    \resizebox{\linewidth}{!}{
    \begin{tabular}{C{0.9cm}C{1.2cm}C{2.4cm}C{0.73cm}C{0.73cm}C{0.73cm}}
    \toprule
Setting & Target set & System & \emph{Pre.} & \emph{Rec.} & $F1$ \\
\cmidrule(lr){1-1} \cmidrule(lr){2-2}  \cmidrule(lr){3-3}  \cmidrule(lr){4-4} \cmidrule(lr){5-5} \cmidrule(lr){6-6}
\multirow{4}{1.5cm}{\rotatebox[origin=c]{90}{\parbox{1.1cm}{\centering T1: \\Perspective \\ relevance}}} & \multirow{4}{*}{ $\perspectivePool$ } & IR & 46.8 & 34.9 & 40.0 \\
& & IR + BERT & 47.3 & 54.8 & \textbf{50.8} \\
\cmidrule(l){3-6}
& & IR + Human & 63.8 & 83.8 & 72.5\\
\midrule
\multirow{4}{1.5cm}{\rotatebox[origin=c]{90}{\parbox{1.1cm}{\centering T2: \\Perspective\\stance}} } & \multirow{4}{*}{\parbox{1.1cm}{$P(c)$}} & Always ``supp.'' & 51.6 & 100.0 & 68.0 \\
& & BERT & 70.5 & 71.1 & \textbf{70.8} \\
\cmidrule(l){3-6}
& & Human & 91.3 & 90.6 & 90.9 \\
\midrule
\multirow{5}{1.5cm}{ \rotatebox[origin=c]{90}{\parbox{1.1cm}{\centering T3: \\Perspective\\equivalence}} } & \multirow{5}{*}{\parbox{1.1cm}{$P(c)^2$}} & Always ``$\neg$equiv.'' & 100.0 & 11.9 & 21.3 \\
& & Always ``equiv.'' & 20.3 & 100.0 & 33.7 \\
& & IR & 36.5 & 36.5 & 36.5 \\
& & BERT & 85.3 & 50.8 & \textbf{63.7} \\
\cmidrule(l){3-6}
& & Human & 87.5 & 80.2 & 83.7\\
\midrule
\multirow{4}{1.5cm}{\rotatebox[origin=c]{90}{\parbox{1.1cm}{\centering T4: \\Evidence \\extraction }} } & \multirow{4}{*}{$\evidencePool$} & IR & 42.2 & 52.5 & 46.8 \\
& & IR + BERT & 69.7 & 46.3 & \textbf{55.7} \\
\cmidrule(l){3-6}
& & IR + Human & 70.8 & 53.1 & 60.7\\
\midrule
\multirow{3}{0.5cm}{\rotatebox[origin=c]{90}{\parbox{1.1cm}{\centering T5: Overall}}} & \multirow{3}{*}{$\perspectivePool, \evidencePool$} & IR & - & - & 12.8 \\
& & IR + BERT & - & - & \textbf{17.5} \\
\cmidrule(l){3-6}
& & IR + Human & - & - & 40.0\\
\bottomrule 
\end{tabular}
    }
    \caption{
    Quality of different baselines on different sub-tasks (Section~\ref{sec:analysis}). 
    All the numbers are in percentage. Top machine baselines are in \textbf{bold}. 
    }
    \label{tab:results}
\end{table}

\section{Discussion}
{


As one of the key consequences of the information revolution, \emph{information pollution} and \emph{over-personalization} have already had detrimental effects on our life. 
In this work, we attempt to facilitate the development of systems that aid in better organization and access to information, with the hope that the access to more diverse information can address over-personalization too~\cite{VZRP14}.

The dataset presented here is not intended to be \emph{exhaustive}, nor does it attempt to reflect a true distribution of the important claims and perspectives in the world, or to associate any of the perspective and identified evidence with levels of expertise and trustworthiness. 
Moreover, it is important to note that when we ask crowd-workers to evaluate the validity of perspectives and evidence, their judgement process can potentially be influenced by their prior beliefs \cite{MarkovitsNa89}. To avoid additional biases introduced in the process of dataset construction, we try to take the least restrictive approach in filtering dataset content beyond the necessary quality assurances. 
For this reason, 
we choose not to explicitly ask annotators to filter contents based on the intention of their creators (e.g. offensive content).




A few algorithmic components were not addressed in this work, although they are important to the complete \emph{perspective discovery and presentation} pipeline.
For instance, one has to first verify that the input to the system is a reasonably well-phrased and an argue-worthy claim. 
And, to construct the pool of perspectives, one has to extract relevant arguments 
\cite{LBHAS14}. 
In a similar vein, since our main focus is the study of the relations between \claim s, \perspective s, and \evidence, we leave out important issues such as their degree of factuality \cite{VlachosRi14} or trustworthiness \cite{PasternackRo14, PasternackRo10a} as separate aspects of problem.}


We hope that some of these challenges and limitations will be addressed in future work.

\section{Conclusion}
\ignore{This work touches upon a class of claims for which answering them requires addressing multiple angles.} 
The importance of this work is three-fold; we define the problem of \emph{substantiated perspective discovery} and characterize language understanding tasks necessary to address this problem.
We combine online resources, web data and crowdsourcing and create a high-quality dataset, in order to drive research on this problem. Finally, we build and evaluate strong baseline supervised systems for this problem. Our hope is that this dataset would bring more attention to this important problem and would speed up the 
progress in this direction. 
 
 There are two aspects that we defer to future work. First, the systems designed here assumed that the input are valid claim sentences. To make use of such systems, one needs to develop mechanisms to recognize valid argumentative structures. In addition, we ignore trustworthiness and credibility issues, important research issues that are addressed in other works.

\section*{Acknowledgments}
The authors would like to thank Jennifer Sheffield, Stephen Mayhew, Shyam Upadhyay, Nitish Gupta and the anonymous reviewers for insightful comments and suggestions. This work was supported in part by a gift from Google and by Contract HR0011-15-2-0025 with the US Defense Advanced Research Projects Agency (DARPA). The views expressed are those of the authors and do not reflect the official policy or position of the Department of Defense or the U.S. Government. 

\bibliography{main}
\bibliographystyle{acl_natbib}

\clearpage

\appendix

\section{Supplemental Material}
\label{sec:supplemental}

\subsection{Statistics}
\label{sec:sup:statistics}
We provide brief statistics on the sources of different content in our dataset in Table~\ref{tab:seed_data}. In particular, this table shows: 
\begin{enumerate}
    \item the size of the data collected from  online debate websites (step 1). 
    \item the size of the data filtered out (step 2a). 
    \item the size of the perspectives added by paraphrases (step 2b). 
    \item the size of the perspective candidates added by web (step 2c). 
\end{enumerate}

\begin{table}[h]
    \small 
    \centering
    \resizebox{\linewidth}{!}{
    \begin{tabular}{ccccc}
    \toprule
    & Website & \# of claims & \# of perspectives & \# of evidences \\
    \cmidrule(lr){2-2}  \cmidrule(lr){3-3}  \cmidrule(lr){4-4} \cmidrule(lr){5-5}
    \parbox[t]{2mm}{\multirow{1}{*}{\rotatebox[origin=c]{90}{after step 1}}} &  {\tt idebate} & 561 & 4136 & 4133 \\
    &  {\tt procon} & 50 & 960 & 953 \\
    &  {\tt debatewise} & 395 & 3039 & 3036 \\
    &  total & 1006 & 8135 & 8122 \\
    \midrule
    \parbox[t]{2mm}{\multirow{2}{*}{\rotatebox[origin=c]{90}{after step 2a}}} &  {\tt idebate} & 537 & 2571 & -- \\
    &  {\tt procon} & 49 & 619 & -- \\
    &  {\tt debatewise} & 361 & 1462 & -- \\
    &  {total} & 947 & 4652 & -- \\
    \midrule
    step 2b &  {paraphrases} & -- & 4507  &  -- \\
    \midrule
    step 2c &  {web perspectives} & -- & 2427  &  -- \\
    \bottomrule 
    
    \end{tabular}
    }
    \caption{
    The dataset statistics (See section~\ref{sec:construction}). }
    \label{tab:seed_data}
\end{table}

\subsection{Measure of agreement}
\label{sec:agreement}
We use the following definition formula in calculation of our measure of agreement. For a fixed subject (problem instance),  
let  $n_j$ represent the number of raters who assigned the given subject to the $j$-th category. The measure of agreement is defined as 
$$
\rho \triangleq \frac{1}{n(n - 1)} \sum_{j=1}^k n_{j} (n_{j} - 1)
$$
where for $n = \sum_{j=1}^k n_{j}$. 
Intuitively, this function measure concentration of values the vector $(n_1, ..., n_k)$. Take the edge cases: 
\begin{itemize}
    \item Values concentrated: $\exists j, n_j = n$ (in other words $\forall i \neq j, n_i = 0$)   $\Rightarrow P = 1.0$. 
    \item Least concentration (uniformly distribution): $n_1 = n_2 = ... = n_k \Rightarrow \rho = 0.0$. 
\end{itemize}

This definition is used in calculation of more extensive agreement measures (e.g, Fleiss' kappa \cite{FleissCo73}). 
There multiple ways of interpreting this formula: 

\begin{itemize}
    \item It indicates how many rater--rater pairs are in agreement, relative to the number of all possible rater--rater pairs. 
    \item One can interpret this measure by a simple combinatorial notions.  Suppose we have sets $A_1,...A_k$ which are pairwise disjunct and for each $j$ let $n_j = \abs{A_j}$. We choose randomly two elements from $A= A_1\cup A_2\cup...\cup A_k$. Then the probability that they are from the same set is the expressed by $\rho$. 

    \item We can write $\rho$ in terms of $\sum_{i=1}^k (n_i-n/k)^2/(n/k)$ which is the conventional \emph{Chi-Square statistic} for testing if the vector of $n_i$ values comes from the all-categories-equally-likely flat multinomial model.  
\end{itemize}

\begin{figure}
    \centering
    \includegraphics[scale=0.25]{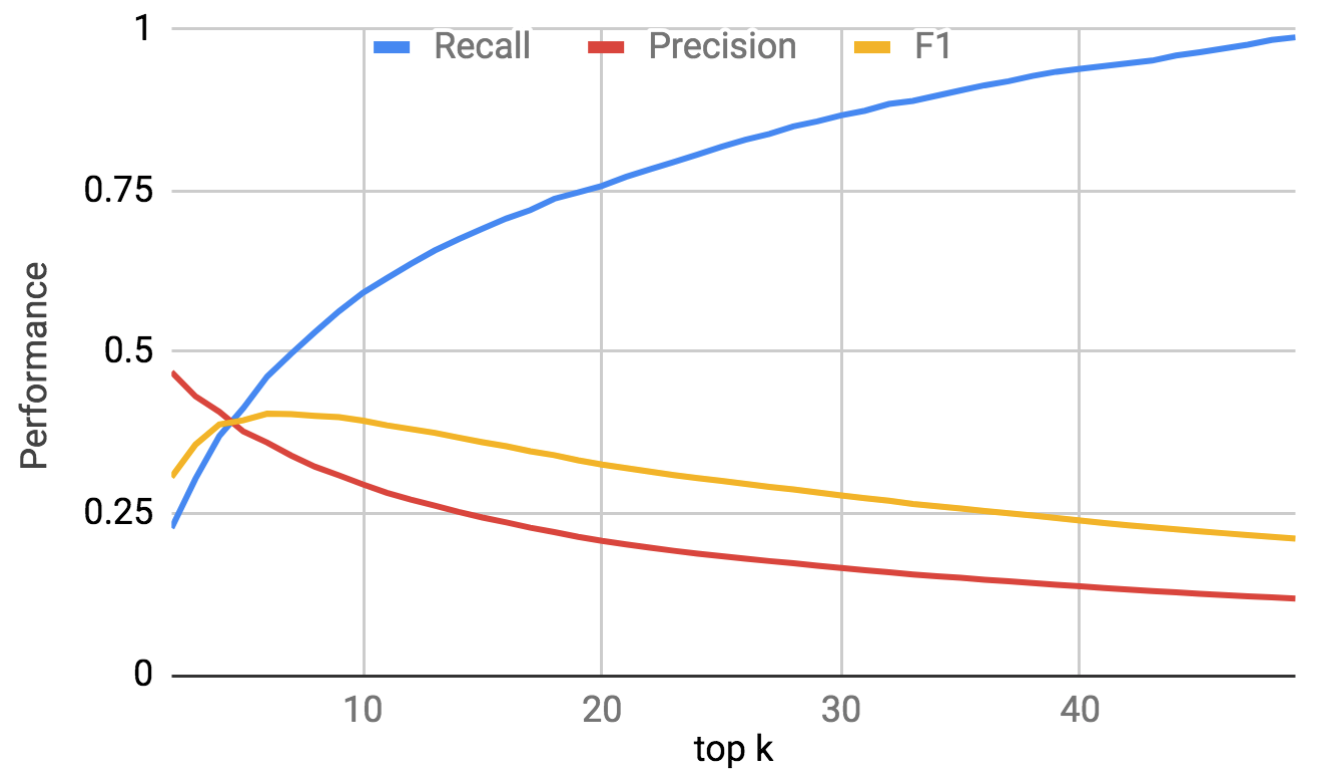}
    \includegraphics[scale=0.25]{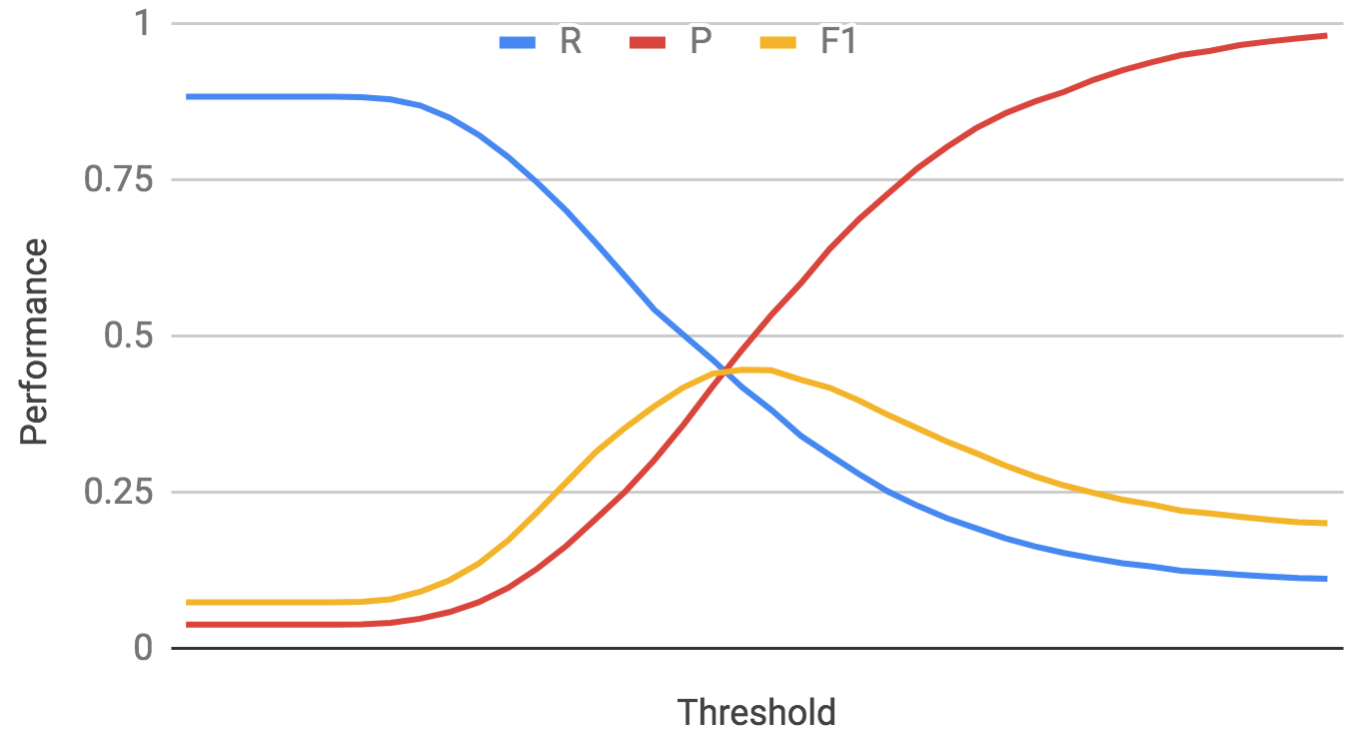}
    \caption{Candidates retrieved from IR baselines vs Precision, Recall, F1, for T1 and T4 respectively. }
    \label{fig:pr-curves}
\end{figure}

\begin{figure*}
    \centering
    \includegraphics[scale=0.45, trim=1.5cm 1cm 0cm 0cm, clip=false]{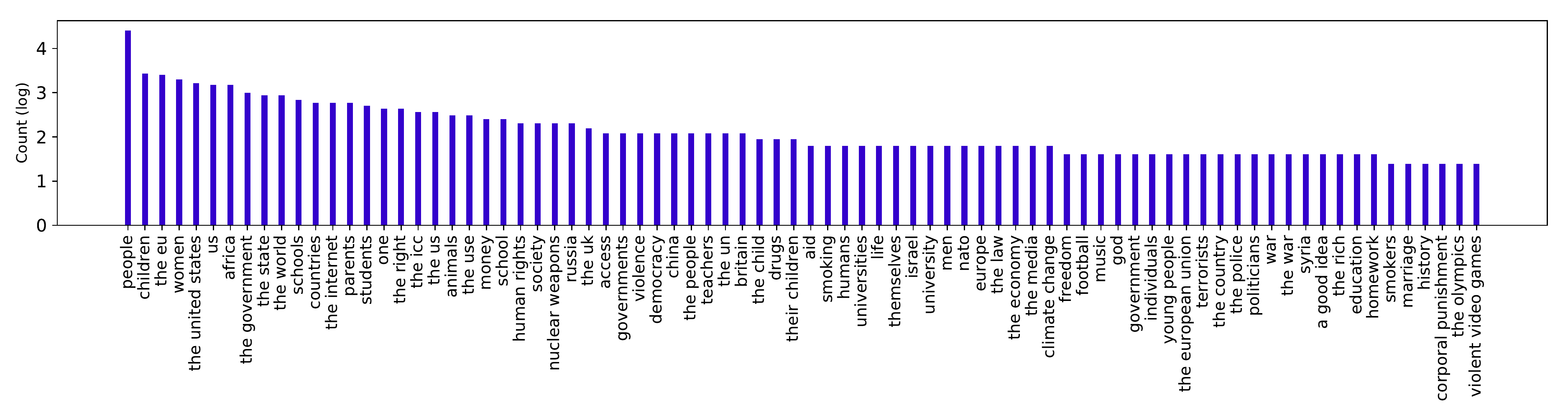}
    \caption{Histogram of popular noun-phrases in our dataset. The $y$-axis shows count in logarithmic scale. }
    \label{fig:noun-phrase-histogram}
\end{figure*}

\begin{figure*}
    \centering
    \includegraphics[scale=0.35,trim=3cm 0cm 2cm 0cm, clip=true]{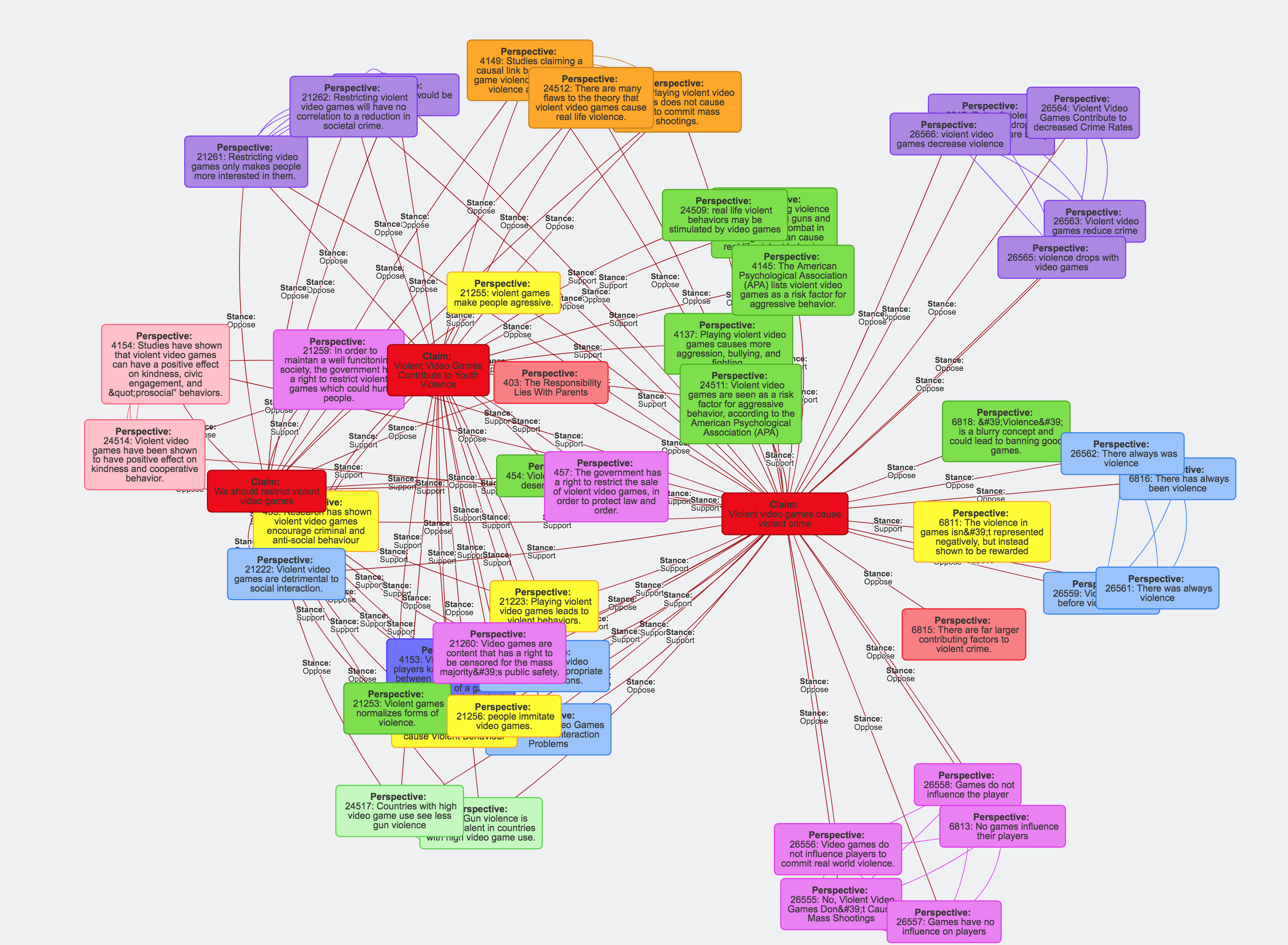}
    \caption{Graph visualization of three related example \emph{claims} (colored in red) in our dataset  with their \emph{perspectives}. Each edge indicates a supporting/opposing relation between a perspective and a claim. }
    \label{fig:violent_games}
\end{figure*}

\subsection{crowdsourcing interfaces}
\label{sec:supp:screenshots}

\begin{figure*}
    \centering
    \frame{\includegraphics[scale=0.33]{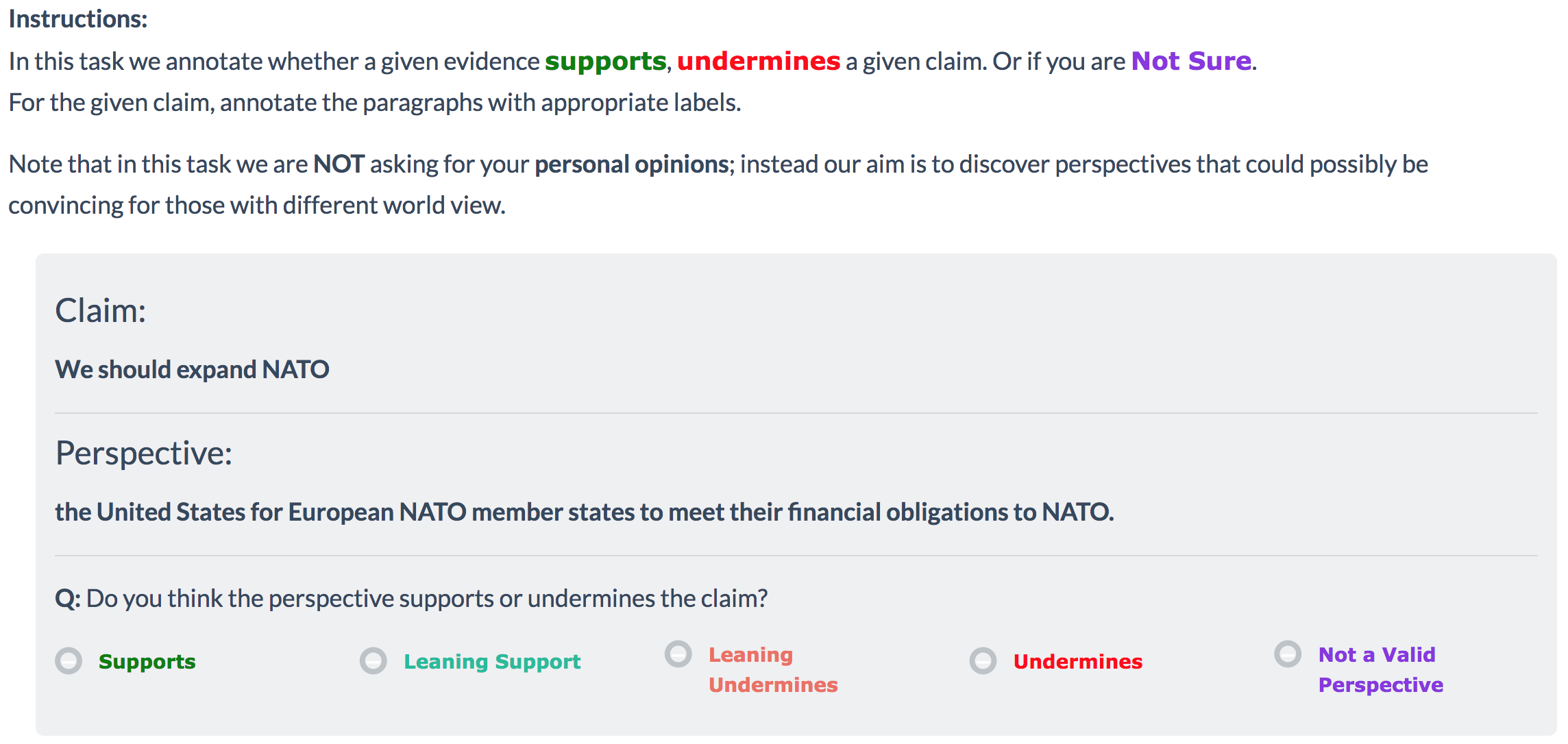}}
    \frame{\includegraphics[scale=0.33]{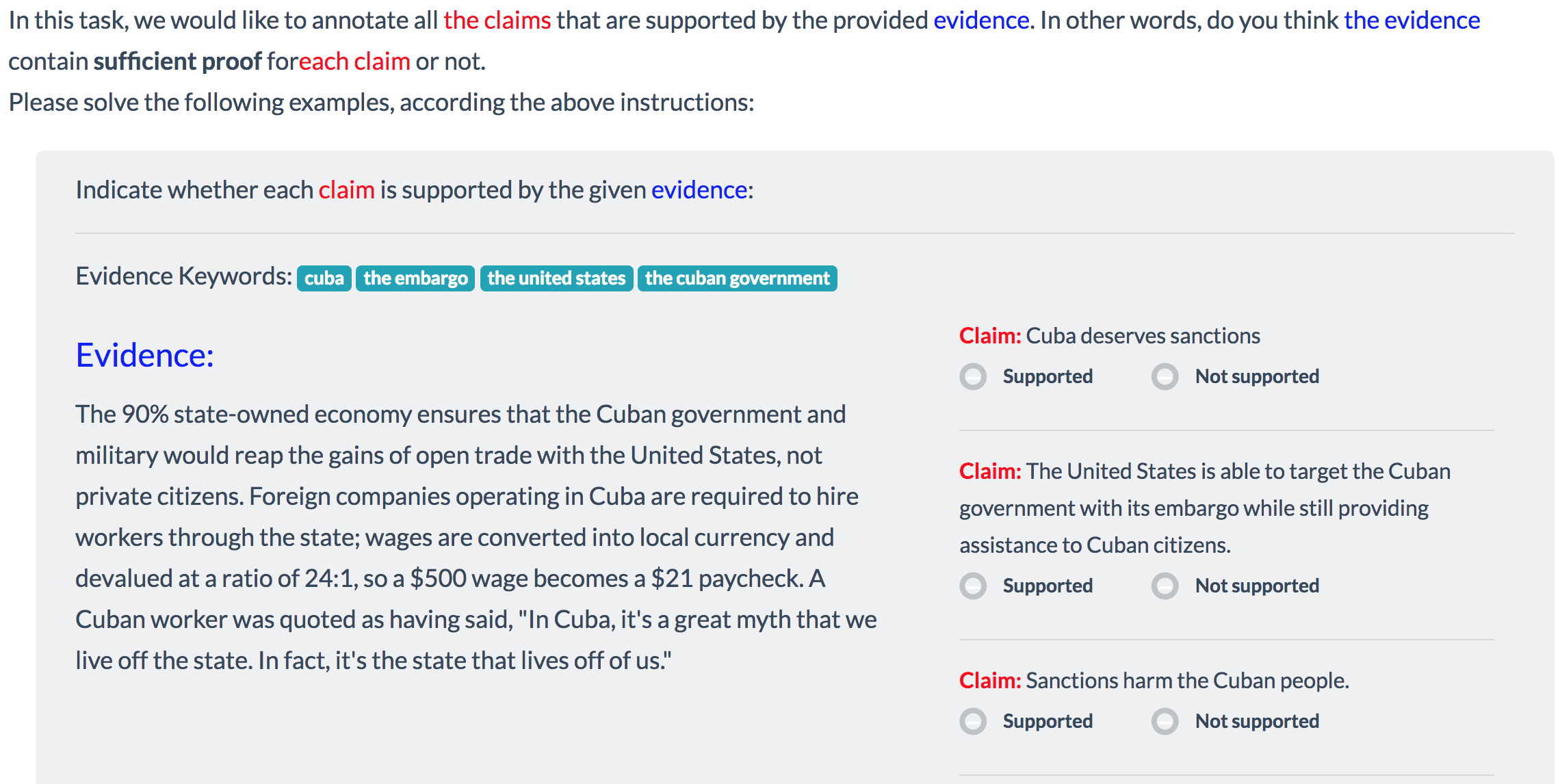}}
    \frame{\includegraphics[scale=0.33]{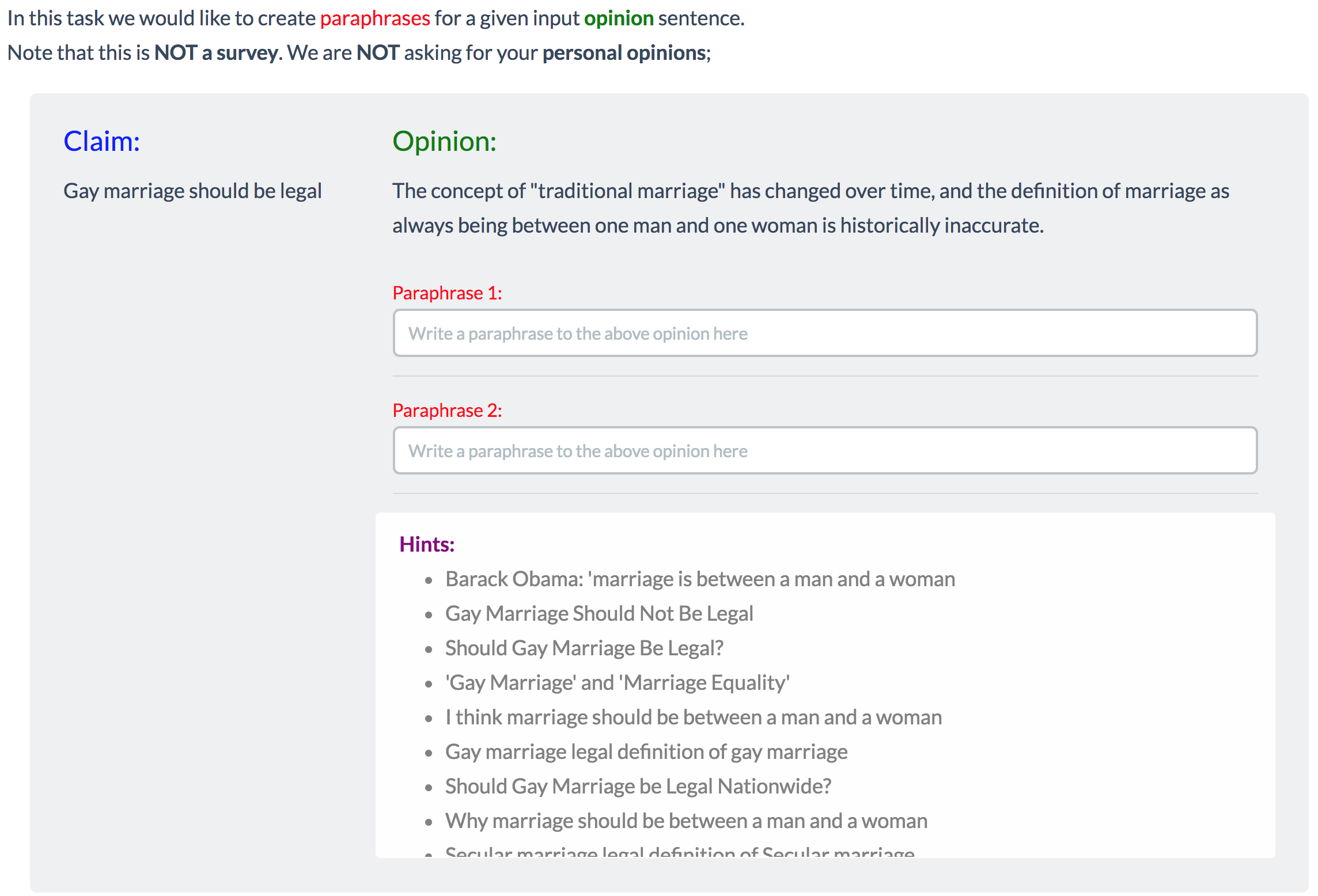}}
    \caption{Interfaces shown to the human annotators. Top: the interface for verification of perspectives (step 2a). Middle: the interface for annotation of evidences (step 3a). Bottom: the interface for generation of perspective paraphrases (step 2b). 
    }
    \label{fig:screenshots1}
\end{figure*}

\begin{figure*}
    \centering
    \frame{\includegraphics[scale=0.33]{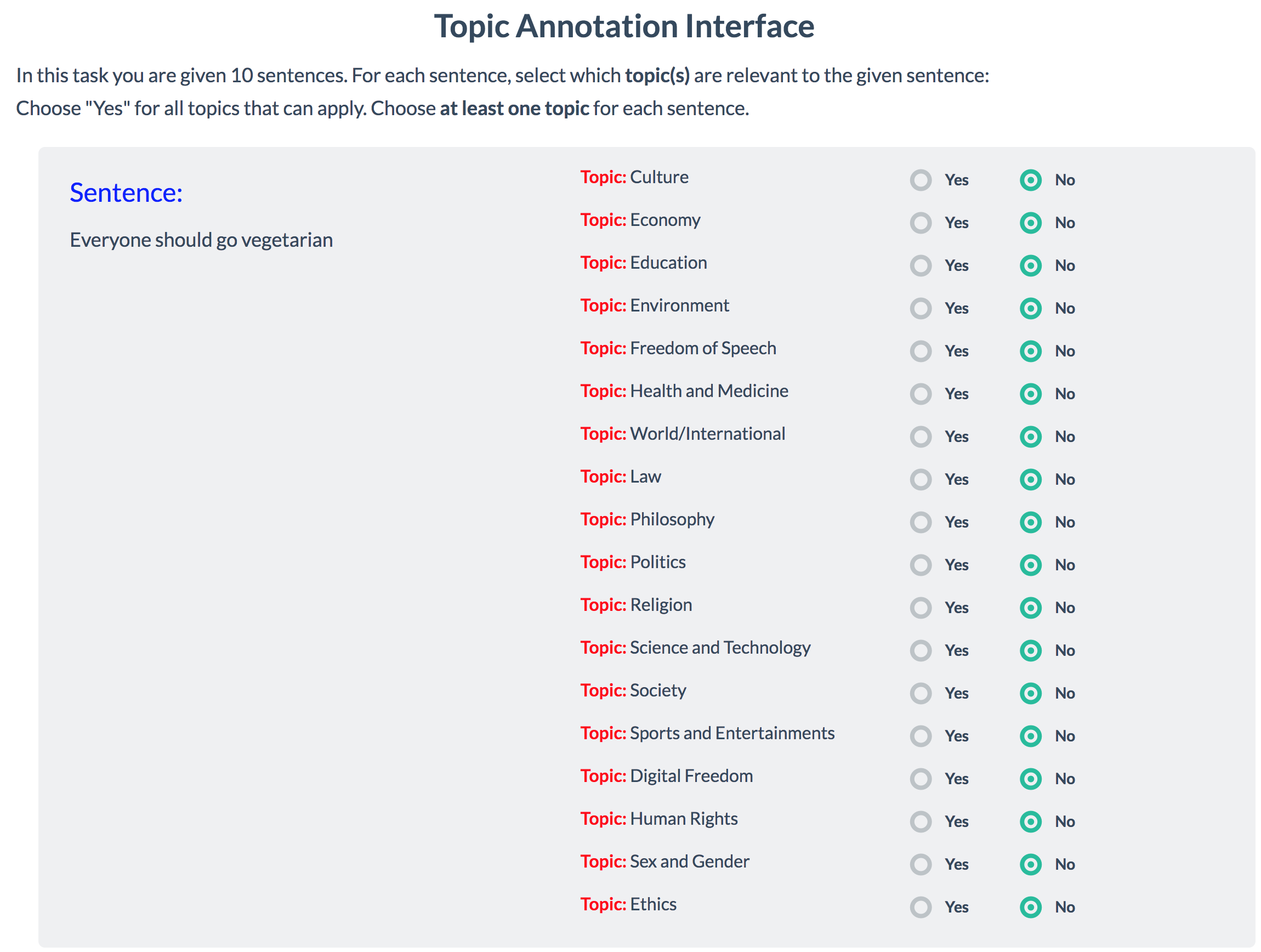}}
    \caption{
    Annotation interface used for topic of claims
    (Section~\ref{sec:statistics}) 
    }
    \label{fig:screenshots1:topic}
\end{figure*}

\end{document}